\documentclass[10pt,twocolumn,letterpaper]{article}

\usepackage{wacv}
\usepackage{times}
\usepackage{epsfig}
\usepackage{graphicx}
\usepackage{amsmath}
\usepackage{amssymb}
\usepackage{array}

\def\wacvPaperID{****} 

\wacvfinalcopy 

\ifwacvfinal
\def\assignedStartPage{1} 
\fi

\ifwacvfinal
\usepackage[breaklinks=true,bookmarks=false]{hyperref}
\else
\usepackage[pagebackref=true,breaklinks=true,colorlinks,bookmarks=false]{hyperref}
\fi

\usepackage{mathtools}

\ifwacvfinal
\else
\fi

\begin{document}

\title{Disentangled Face Identity Representations \\ for joint 3D Face Recognition and Expression Neutralisation}

\author{Anis Kacem\\
SnT, University of Luxembourg\\

\and

Kseniya Cherenkova\\
SnT, University of Luxembourg, Artec3D\\
\and
Djamila Aouada\\
SnT, University of Luxembourg\\
}

\maketitle

\begin{abstract}
  In this paper, we propose a new deep learning based approach for disentangling face identity representations from expressive 3D faces. Given a 3D face, our approach not only extracts a disentangled identity representation, but also generates a realistic 3D face with a neutral expression while predicting its identity. The proposed network consists of three components; (1) a Graph Convolutional Autoencoder (GCA) to encode the 3D faces into latent representations, (2) a Generative Adversarial Network (GAN) that translates the latent representations of expressive faces into those of neutral faces, (3) and an identity recognition sub-network taking advantage of the neutralized latent representations for 3D face recognition. The whole network is trained in an end-to-end manner. Experiments are conducted on three publicly available datasets showing the effectiveness of the proposed approach.
\end{abstract}

\section{Introduction}

Automatic recognition of faces is a non-intrusive technology that, if performed accurately, can open the door to many innovative applications and revolutionize the interactions of humans with infrastructures and services. With the recent advances in acquisition systems, impressive face recognition systems have been deployed in real-world applications \cite{drira20133d,li2015towards,lei2014efficient,parkhi2015deep,liu2017sphereface,schroff2015facenet,faceid}. In particular, 3D sensors played an important role in making these advances thanks to the rich geometrical information and their robustness to multiple real-world constraints (e.g., pose variations, illumination changes, etc) \cite{petrovska2009guide}. 

Another important constraint in real-world scenarios consists of recognizing faces under different facial deformations usually termed as facial expressions \cite{drira20133d,amberg2008expression}. Indeed, the captured information when acquiring faces encodes different attributes, e.g., identity, expression, pose, etc \cite{blanz1999morphable,amberg2008expression,jiang2019disentangled}. Recognizing faces usually requires decoupling identity attributes from other attributes. Following this direction, multiple works tried to build 3D face parametric models disentangling identity attributes from expression ones. In particular, 3D Morphable Models (3DMM) have been widely used in different forms to achieve this goal \cite{blanz1999morphable,amberg2008expression,booth20163d}. Such models usually assume linear combination of the different attributes which limits their disentanglement abilities due to the natural non-linearity of face shape variations. Aware of such limitations, recent works proposed non-linear models to model 3D face shapes~\cite{jiang2019disentangled,cheng2019meshgan,bouritsas2019neural,ranjan2018generating}. \cite{ranjan2018generating} proposed a Graph Convolutional Autoencoder (GCA) making use of spectral graph convolutions~\cite{bruna2013spectral} to encode 3D face shapes into non-linear latent representations. \cite{cheng2019meshgan} presented an intrinsic Generative Adversarial Network (GAN) architecture named as MeshGAN operating directly on 3D face meshes using a similar strategy as~\cite{ranjan2018generating}. Their proposed method allows the generation of new identities and expressions. Similarly, the authors in~\cite{jiang2019disentangled} decoupled identity and expression representations by building two separate branches of networks, one for identity and the other for expression. Their proposed networks were based on spectral graph convolutions and were applied on deformation representations computed on the face meshes in advance. In order to control the distribution in latent space, they used a variational strategy when training each branch.
Taking another direction, \cite{bouritsas2019neural} introduced spiral convolutions on meshes and used them within a GAN-based architecture to synthesise face meshes.

\begin{figure*}[h]
  \includegraphics[width=\textwidth]{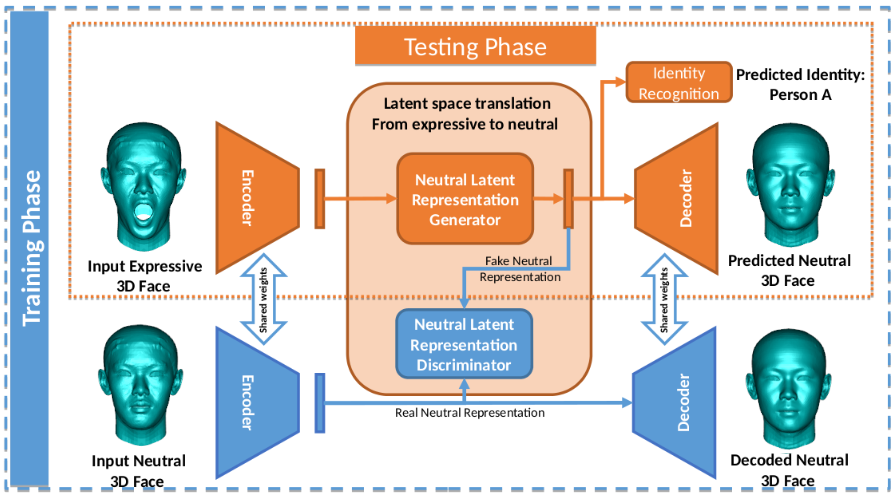}
  \caption{Approach Overview. In training phase: pairs of expressive and neutral 3D faces of the same identity are collected. These pairs are fed to a two-branch autoencoder sharing the same weights. A GAN-based module consisting of a generator and a discriminator is introduced in the latent space to learn a valid translation from expressive to neutral representation. An identity recognition module is adopted to learn better identity representations.
  In testing phase: An expressive 3D face is given as input to the network. The encoder encodes the 3D face into a latent representation. Then, the generator of the GAN module translates this latent representation into neutral one which is then decoded into a neutral 3D face. The output features of the face recognition module can be also used to conduct face recognition.  
  Best viewed in color. }
  \label{fig:overview}
\end{figure*}

The main focus of this paper is to effectively disentangle identity attributes from expression ones and use them within a face recognition scenario. Accordingly, we learn a non-linear mapping translating expressive 3D faces to neutral ones. This is achieved at the level of latent representations obtained using a Graph Convolutional Autoencoder (GCA)~\cite{ranjan2018generating}. The latent space translation is learned using a Generative Adversarial Network (GAN) applied on pairs of expressive and neutral latent  representations~\cite{isola2017image}. Moreover, additional modules and constraints are proposed to better preserve identity features when neutralizing expressions. Figure~\ref{fig:overview} shows an overview of the proposed approach. The main contributions of this paper are:

\begin{itemize}
    \item An end-to-end network allowing joint face expression neutralization and face recognition. Given an expressive 3D face, the proposed network can also synthesise the corresponding realistic neutral 3D face;  
    \item A novel data-driven disentangled identity representation obtained by a realistic translation from expressive faces to neutral ones;
    \item An experimental validation of the proposed approach on three publicly available datasets.
\end{itemize}

The rest of the paper is organized as follows: 
Section~\ref{sect:app} describes the proposed approach. In Section~\ref{sect:inference}, we provide the network architecture and details on the training. The experimental evaluation is reported in Section~\ref{sect:experiments}. Finally, Section~\ref{sect:conclusion} concludes the paper and draws some perspectives.






\section{Proposed Approach}
\label{sect:app}



Let us consider a set of $m$ face meshes $\{F_j\}^m_{j=1}$. A face mesh is defined as a graph structure, $F = (V, A)$. Here, $V=\left[v_1,v_2,\dots,v_n\right]$ denotes the 3D coordinates of $n$ vertices and $A$ represents an $n \times n$ adjacency matrix with $A_{ij}=1$ if an edge connection exists between vertex $v_i$ and $v_j$, and $A_{ij}=0$ otherwise. 

A set of pairs of expressive and neutral 3D faces are collected such that each pair $(F^{e}_i,F^{n}_i)$ involves an expressive 3D face $F^{e}_i$ and its neutral counterpart $F^{n}_i$ (\emph{i.e.}, a neutral 3D face of the same identity). The main objective of our approach is to translate the expressive face $F^{e}_i$ into a neutral face $\Tilde{F^{n}_i}$ that approximates well the ground truth neutral face $F^{n}_i$, while allowing the prediction of an identity code $\Tilde{y_i}$ that should match the ground truth identity label $y_i$ of both $F^{e}_i$ and $F^{n}_i$.

\subsection{Graph Convolutional Autoencoder}
\label{sect:GCA}
The first step to our approach is to to embed the input 3D face meshes into a low-dimensional latent space that encodes the expression and identity attributes of the input face meshes. To do so, we use a mesh-based autoencoder consisting of an encoder $Enc$ and a decoder $Dec$. The encoder $Enc$ maps the input face pair $(F^{e}_i,F^{n}_i)$ into latent representations $(z_i^{e},z_i^{n})$. The decoder $Dec$ directly maps back the neutral latent representation $z_i^{n}$ into a neutral face mesh $\Tilde{F^{n}_i}$ and also maps the neutralised version of the expressive latent representation $z_i^{e}$ into the same neutral face mesh $\Tilde{F^{n}_i}$. The neutralization of the latent representation will be explained in the next section. From an implementation point of view, the proposed auto-encoder can be seen as two branches of Encoder-Decoder sharing the same weights. 

Due to the graph structure of the input face meshes, standard convolutions cannot be used in $Enc$ and $Dec$. As a solution, we use the spectral graph convolutions proposed in~\cite{bruna2013spectral}. In what follows, we recall the background related to these convolutions.

\textbf{Spectral Graph Convolutions:} These convolutions are defined on the graph Fourier transform domain. The Fourier transform of the vertices $V$ is given by $V_w = U^t V$, where $U \in \mathbb{R}^{n \times n}$ whose columns are the orthogonal eigenvectors of the associated Laplacian matrix $L$, while the inverse Fourier transform is defined as $V = U V_w$. The associated Laplacian to a graph $F$ is defined as $L=\Lambda-A$, where $\Lambda$ is a diagonal matrix representing the degree of each vertex in $V$ with diagonal elements $\Lambda_{ii}=\sum_{j=1}^n A_{ij}$.  The convolution operator in Fourier space can be defined as a Hadamard product of the Fourier transforms of the graph and the filter. For efficiency matters, it is common to use a recursive Chebyshev polynomial of order $P$ \cite{ranjan2018generating,defferrard2016convolutional} to approximate a kernel $k_\theta(L) = \sum_{p=0}^{P-1} \theta_{p}T_p(\hat{L})$, where $\hat{L}=2L/e_{max} - I_n$ is the scaled Laplacian matrix, $e_{max}$  maximal eigenvalue of $L$, $\theta \in \mathbb{R}^{P}$ is a vector of Chebyshev coefficients that should be learned, and $T_p \in \mathbb{R}^{n \times n}$ is the Chebyshev polynomial of order $p$ that can be computed recursively~\cite{defferrard2016convolutional}. The spectral convolution is finally defined as, 
\begin{equation}
    c_j = \sum_{i=1}^{n_f} k_{\theta_{i,j}}(L) V_{:,i} \in \mathbb{R}^n,
\end{equation}

\noindent where $c_j$ is the $j$-th feature of the output $c \in \mathbb{R}^{n \times n_{of}}$ ($n_{of}$ is the number of output features), $V_{:,i}$ is the $i$-th feature of the input $V \in \mathbb{R}^{n \times n_{f}}$ ($n_{f}$ is the number of input features), and $\theta_{i,j} \in \mathbb{R}^P$ are the Chebyshev trainable coefficients. For more details about spectral graph convolutions, readers are refereed to~\cite{defferrard2016convolutional,ranjan2018generating}.

For notation simplicity, we will use in the following $z_i^{e} = Enc(F^{e}_i)$ to denote the expressive latent representation and $z_i^{n} = Enc(F^{n}_i)$ to represent the neutral latent representation.

\subsection{Latent Space Translation}
\label{sect:l2l}
Given a set of pairs of latent representations $(z_i^{e},z_i^{n})$ learned from the expressive and neutral face meshes $(F^{e}_i,F^{n}_i)$, we aim to find a mapping function $\mathcal{G}: z_i^{e} \rightarrow z_i^{n}$ that translates the expressive latent representation into its corresponding neutral latent representation. To approximate this function we use a neural network $\mathcal{G}$ on the expressive latent representation $z_i^{e}$ and constrain its output to be close to the neutral latent representation $z_i^{n}$ by minimizing the L1 distance: 

\begin{equation}
    \mathcal{L}_{L1}= \left\lVert \mathcal{G}(z_i^{e})- z_i^{n}\right\rVert_{1} .
\label{eq:rec_l}
\end{equation}

For a more realistic translation, we further constrain the distribution of expressive latent representations to match that of neutral latent representations. This is achieved by employing a conditional Generative Adversarial Network (cGAN)~\cite{mirza2014conditional,isola2017image} on these latent representations. The generator of this cGAN is given by the network $\mathcal{G}$ that takes the expressive latent representations as input and tries to generate the corresponding neutral latent representations. The discriminator $\mathcal{D}$ is defined as another network that considers the neutral latent representations $\mathcal{G}(z_i^{e})$ as fake examples, while taking the neutral latent representations $z_i^{n}$ as real examples. Furthermore, the discriminator $\mathcal{D}$ is conditioned using the expressive latent representation $z_i^{e}$ as proposed in~\cite{isola2017image} for image-to-image translation. This condition is established by concatenating the condition $z_i^{e}$ to the inputs of the discriminator (\emph{i.e.}, $z_i^{n}$ or $\mathcal{G}(z_i^{e})$). The adversarial loss to be minimized is then defined as,

\begin{equation}
\begin{multlined}
\mathcal{L}_{GAN}= \mathbb{E}_{z_i^{e},z_i^{n}}\left[ \log \mathcal{D}(z_i^{n},z_i^{e})\right]  + \\
 \mathbb{E}_{z_i^{e},z_i^{n}}\left[\log(1 - \mathcal{D}(\mathcal{G}(z_i^{e}),z_i^{e}))   \right].
\end{multlined}
\label{eq:adv}
\end{equation}

The total loss function for latent space translation is given by, 

\begin{equation}
\mathcal{L}_{l2l}=  \lambda_{L1} \mathcal{L}_{L1} +  \lambda_{GAN} \mathcal{L}_{GAN} ,
\label{eq:l2l}
\end{equation}

\noindent where $\lambda_{L1}$ and $\lambda_{GAN}$ denote the weights of $\mathcal{L}_{L1}$ and $\mathcal{L}_{GAN}$, respectively.  


\subsection{Face Recognition Module}
\label{sect:frec}

In order to conduct face recognition, we take advantage of the generated neutral latent representation for an expression-robust face recognition. Specifically, we use a neural network $\mathcal{R}$ on the generated neutral latent representation $\mathcal{G}(z_i^{e})$ to predict the identity $\Tilde{y_i} = \mathcal{R}(\mathcal{G}(z_i^{e}))$ of the expressive face $F^{e}_i$. The same network is also used on the real neutral latent representation $z_i^{n}$ to obtain a prediction $\Tilde{y_i} =\mathcal{R}(z_i^{n})$. 
The standard cross-entropy loss is used for training over the identities as follows,

\begin{equation}
\mathcal{L}_{id} = - \sum^{s}_{j=1} y^j_i \log \Tilde{y}^j_i , 
\label{eq:id}
\end{equation}

\noindent where $s$ denotes the number of subjects (\emph{i.e.}, classes) and $y^j_i$ is is the $j$-th value of the ground truth label $y_i$. 

It is important to mention that the face recognition module is not only used for predicting identities but also for a better preservation of the identity information when neutralizing the expression. We experimentally validate this claim in Section~\ref{sect:exp_en}.

\section{Final Objective and Network Architecture}
\label{sect:inference}
In addition to the loss functions of latent space translation and face recognition module defined in Eq.~\ref{eq:l2l} and Eq.~\ref{eq:id}, respectively, we further define two loss functions at the end of the decoder $Dec$. The first one enforces the autoencoder to reconstruct an input neutral face $F^{n}_i$, while the second constrains the whole network to reconstruct, starting from an expressive face $F^{e}_i$, an output neutral face that is close to the input neutral face $F^{n}_i$. These reconstruction losses are formulated using an $L_1$ loss as follows, 

\begin{equation}
\begin{multlined}
\mathcal{L}_{rec} = \left\lVert Dec( Enc (F_i^{n}))- F_i^{n}\right\rVert_{1}  + \\
\left\lVert Dec( \mathcal{G}(Enc (F_i^{e})))- F_i^{n}\right\rVert_{1} ,
\end{multlined}
\label{eq:rec}
\end{equation}

The final objective to be minimised is finally defined as, 

\begin{equation}
\mathcal{L}_{tot}=  \lambda_{l2l} \mathcal{L}_{l2l} +  \lambda_{id} \mathcal{L}_{id} + \lambda_{rec} \mathcal{L}_{rec} , 
\label{eq:tot}
\end{equation}

\noindent where $\lambda_{l2l}$, $\lambda_{id}$, and $\lambda_{rec}$ represent the weights of the loss functions  $\mathcal{L}_{l2l}$, $\mathcal{L}_{id}$, and $\mathcal{L}_{rec}$, respectively. 

\textbf{Inference details:} As mentioned in Section~\ref{sect:app}, it is crucial for the training phase of our approach to collect pairs of expressive face meshes and their corresponding neutral faces. These pairs are used to train the latent space translator introduced in Section~\ref{sect:l2l}. During testing phase, the only required input is an expressive face mesh $F^{e}_i$ that is encoded using $Enc$ to obtain an expressive latent representation $z_i^{e}$. The latter is then fed to the generator $\mathcal{G}$ to generate a neutral latent representation $z_i^{n}$. The generated neutral representation is finally passed to the face recognition module $\mathcal{R}$ to predict the identity if the input expressive face and also decoded to produce a neutralized face mesh. 

\begin{figure*}[h]
  \includegraphics[width=\textwidth]{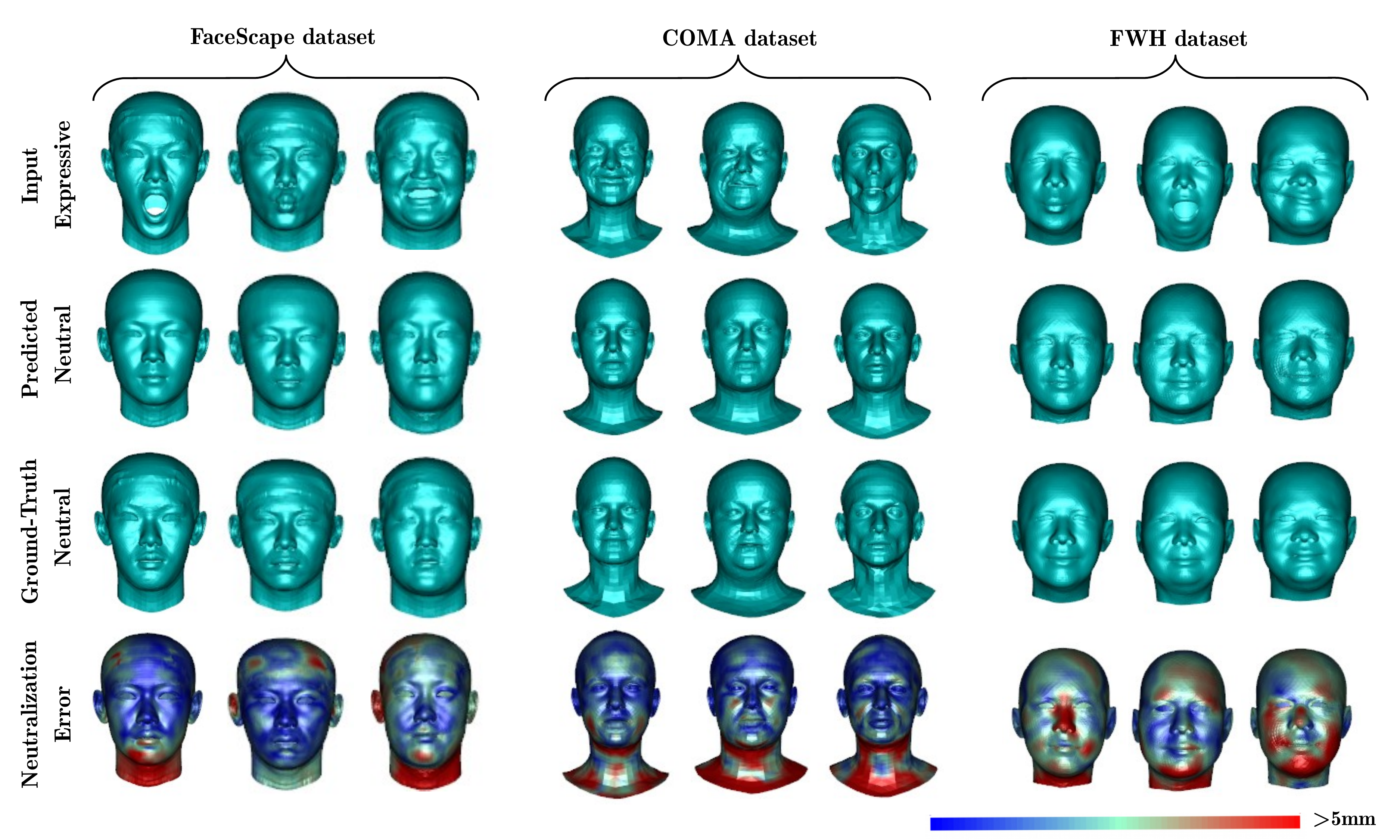}
  \caption{Face expression neutralization results on FaceScape, COMA, and FWH datasets. The first row shows different input expressive 3D faces, the second row presents the predicted neutralized 3D faces, the third row depicts the corresponding ground-truth neutral 3D faces, and the last row depicts the errors of the neutralization computed using per-vertex Euclidean distances between predicted neutral (second row) and ground-truth neutral (third row) 3D faces. Best viewed in color.}
  \label{fig:results}
\end{figure*}
\textbf{Network architecture:} The encoder $Enc$ has four Chebyshev convolutional layers. The first three layers have $16$ output features per vertex and the last layer has $32$ output features. The output of the last convolutional layer is fed into a (Fully Connected) FC layer to obtain a vector latent representation of size $25$. 
The decoder $Dec$ consists of a FC layer that transforms back the latent representation to a graph representation of $n$ vertices with $32$ features per vertex. The output of this FC is then passed to four Chebyshev convolutional layers with a number of output features $32$, $16$, $16$, and $3$, respectively. Each Chebyshev convolutional layer in $Enc$ and $Dec$  is approximated using $P=6$ polynomials and is followed by a biased ReLU~\cite{glorot2011deep}. Regarding the latent representation generator $\mathcal{G}$, we use a succession of four FC layers with a number of output features $100$, $200$, $50$, and $25$, respectively. Each layer is followed by a ReLU activation. A similar architecture is employed for the discriminator $\mathcal{D}$ which consists of four FC layers of output size $100$, $200$, $50$, and $1$, respectively. The first three FC layers are followed by Leaky ReLU activations and the last layer is followed by a Sigmoid activation. Finally, the face recognition module $\mathcal{R}$ consists of two FC layers, the first one has an output size $100$ and is followed by a ReLU activation, and the second one has an output size equal to the number of training subjects and is followed by a Softmax layer. 

In all our experiments we empirically set the weights of the loss function terms as follows: $\lambda_{l2l} = 0.5$, $\lambda_{L1} = 0.4$, $\lambda_{GAN} = 1 $,  $\lambda_{id} = 0.05 $, and $\lambda_{rec} = 2$



\section{Experiments}
\label{sect:experiments}

We experimentally validate the proposed approach on three publicly available datasets. On one hand, we evaluate the face neutralization and compare it with some state-of-art methods. On the other hand, we conduct face recognition based on the disentagled identity representations and compare our results with baselines. The used datasets are the following:


\textbf{COMA dataset~\cite{ranjan2018generating}:} is a dataset designed for dynamic 3D face analysis. It contains more than $20$k 3D faces collected from $12$ identities. Each subject performed $12$ facial expressions showing different challenges.  

\textbf{FaceWareHouse (FWH) dataset~\cite{cao2013facewarehouse}:} is a 3D face shape dataset including $47$ expressions (including neutral) performed by $150$ different identities. The intensity of the expressions involved in this dataset varies from very weak to very strong expressions. 

\textbf{FaceScape dataset~\cite{yang2020facescape}:} consists of a dataset of $938$ subjects, each one performed $21$ expressions inspired from the FWH dataset (including neutral).


\subsection{Face Expression Neutralization}
\label{sect:exp_en}

We evaluate the effectiveness of the expression neutralization by comparing the neutralized expressive face to its ground-truth neutral counterpart. 

\textbf{Experimental protocol:} We split the three datasets into training and testing set following a cross-subject protocol. $70$\% of the subjects were used for training and the rest for testing. The fitted base models given by the considered datasets have been employed since the proposed approach assumes registered 3D face shapes. The pairs of expressive and neutral 3D faces are directly given by the FWH and FaceScape datasets since both involve static 3D face shapes with different expressions (including neutral). In the dynamic COMA dataset, we take the first frame of each 3D face sequence as neutral face and select $12$ expressive faces from other frames.

\textbf{Results and discussions:} In Figure~\ref{fig:results}, we report qualitative results of the face expression neutralization process. In particular, given some input expressive faces selected from the three considered datasets, we present the predicted neutralized faces and compare them to the corresponding ground-truth neutral faces using a per-vertex Euclidean distance. Overall, our approach can perfectly neutralize weak and strong expressions on all the datasets. However, we can observe that the neutralization on FaceScape dataset preserves better the identity features. For example, the facial wrinkles were not properly reconstructed on some examples of COMA and FWH datasets. This can be explained by the larger number of identities included in the FaceScape dataset compared to FWH and COMA datasets. This suggests that the generalization ability of the proposed network to unseen subjects can be impacted by the number of identities considered in the training. 

In Table~\ref{tab:MSE}, we quantitatively evaluate the expression neutralization results by computing the mean errors on the testing set of all datasets between predicted neutral 3D faces and ground-truth ones. In accordance with the observed qualitative results, the lowest error was recorded on FaceScape dataset reaching a mean error of 2.02~millimeters (mm). The highest error was obtained the FWH dataset corresponding to 3.07~mm. On COMA dataset, we obtained an error of 2.73~mm.  

We also compare our results to an adapted version of the Graph Convolutional Autoencoder (GCA) method \cite{ranjan2018generating}. In particular, we trained the GCA network by considering the prepared pairs of expressive and neutral faces as input and ground-truth faces, respectively. The mean errors of the adapted GCA were clearly outperformed by our approach on all the datasets. A further comparison has been conducted with the Disentagled Representation Learning (DRL) approach ~\cite{jiang2019disentangled}. Specifically, we considered the output of the identity branch of the DRL network as predicted neutral face and evaluated the mean errors following the same protocol as our approach. In Table~\ref{tab:MSE}, we can note that the DRL approach was more accurate than ours by more than 1~mm on the FWH dataset. It is important to note that DRL method relies on an intermediate representation called \textit{Deformation Representation} (DR) which needs to be computed before training and/or inference. Another important step in DRL approach is the data augmentation performed on the DR representations to increase the number of identities in the dataset. In particular, 10k faces were added on top of the original FWH dataset used in our approach.

Finally, we conduct an ablation study of the two loss function terms $\mathcal{L}_{l2l}$ and $\mathcal{L}_{id}$ defined in Eq.~\ref{eq:l2l} and Eq.~\ref{eq:id}, respectively. In Table.~\ref{tab:MSE}, we can observe that dropping one of these terms results in a lower performance showing their importance in the training.

\begin{table}[h]
\centering
\begin{tabular}{|c|c|c|c|}
\hline
\textbf{Method}                                      & \textbf{COMA } & \textbf{FWH } & \textbf{FaceScape } \\ \hline
GCA \cite{ranjan2018generating}                         &       3.28        &    3.56          &             2.88       \\ \hline
DLR \cite{jiang2019disentangled}  &     -          &   1.83             &             -     \\ \hline
Ours w/o $\mathcal{L}_{l2l}$                         &          3.45     &           3.92   &            3.29        \\ \hline
Ours w/o $\mathcal{L}_{id}$                          &           3.06    &          3.35    &                2.56    \\ \hline
\textbf{Ours}                                        &             2.73  &        3.07     &        2.02            \\ \hline
\end{tabular}
\caption{Comparison of mean error of expression neutralization on different datasets. All errors are in millimeters.}
\label{tab:MSE}
\end{table}

\subsection{Face Recognition}
\label{sect:exp_frec}

A direct application of the disentangled identity representation obtained using our method consists of face recognition. As mentioned in Section~\ref{sect:frec}, we introduced a face recognition module that improved the face neutralization by enhancing identity preservation as shown in Section~\ref{sect:exp_en}. Such module can also be used to predict the identity of an input face mesh under different facial expressions. To validate this claim, we conducted face identification experiments based on the embedding of the face recognition module $\mathcal{R}$. In particular, we consider the identity features obtained by the last FC layer of the face recognition module (before applying softmax) as an identity representation of an input 3D face. These representations are then compared using cosine similarity measure to achieve face identification. 

\textbf{Experimental protocol:} For face identification, only FWH and FaceScape datasets were retained since COMA dataset consists of only $12$ subjects making not useful to evaluate face identification. As done for expression neutralization experiments, the two considered datasets were split into training and testing sets following a cross-subject protocol. The training of the whole network was achieved on the training set and the testing set of unseen subjects was considered to evaluate face identification. We considered neutral 3D faces of the testing set as gallery and the remaining expressions as probe for the two datasets. The evaluation metric was the rank-1 accuracy.

\begin{table}[h]
\centering
\begin{tabular}{|c|c|c|}
\hline
\textbf{Method}                                      & \textbf{FWH (\%)} & \textbf{FaceScape (\%)} \\ \hline
GCA \cite{ranjan2018generating}                                     &  90.48            &               95.64     \\ \hline
Ours w/o $\mathcal{L}_{l2l}$                                       &    88.75          &         97.83           \\ \hline
\textbf{Ours}                                                   &          92.34    &            99.88        \\ \hline
\end{tabular}
\caption{Comparison of rank-1 accuracies of face identification on FWH and FaceScape datasets. }
\label{tab:frec}
\end{table}

\textbf{Results and discussions:} In Table~\ref{tab:frec}, we report the rank-1 accuracies obtained using our method on FaceScape and FWH datasets. On the FaceScape dataset which includes a large number of identities, we achieved a rank-1 accuracy of 99.88\%. A lower performance consisting of 92.34\% was obtained on the FWH dataset which involves lower number of identities than the FaceScape dataset. As noticed for expression neutralization in Section~\ref{sect:exp_en}, face identification results suggest that the performance of the proposed method is better when a large number of identities are involved in the training. Similarly to expression neutralization experiments in Section~\ref{sect:exp_en}, we compare our face identification results with the adapted GCA method \cite{ranjan2018generating}. The considered identity features to conduct face identification with GCA were the latent representations of the input expressive faces (i.e., bottleneck of the autoencoder). Our approach outperformed the GCA method by around 4\% and 2\% on FaceScape and FHW datasets, respectively. 

We also evaluate our results when discarding the $\mathcal{L}_{l2l}$ loss term. This means that expression neutralization is only ensured by the reconstruction losses defined in Eq.~\ref{eq:rec}. Unsurprisingly, face identification accuracies were lower than those achieved by the retained approach on the two considered datasets.

\section{Conclusion And Future Works}
\label{sect:conclusion}

In this paper, we proposed a neural network method for learning disentangled identity representations. Given an input expressive 3D face, the proposed network neutralizes the facial expression and also provides relevant identity features for face identification. This is achieved by translating expressive latent space representations of 3D faces obtained using spectral graph convolutions to neutral ones. The obtained results of face expression neutralization and face identification on three publicly available datasets showed the effectiveness of the proposed approach. However, some limitations were also noticed when the number of identities is not sufficient which resulted in less generalization abilities to unseen subjects. A data augmentation step increasing face shape variations before training can be introduced in the future to overcome this problem. As additional future work, the collection of expressive and neutral pairs can be avoided by considering an unpaired translation of the latent space such as in~\cite{zhu2017unpaired}.

\section*{Acknowledgement}
This work was funded by the National Research Fund (FNR), Luxembourg, under the project reference CPPP17/IS/11643091/IDform/Aouada

\end{document}